\documentclass[lnbip]{llncs}
\usepackage{makeidx}  
\usepackage{graphicx}
\usepackage{url}
\usepackage{floatrow}%

\begin{document}
\mainmatter              
\title{Civique:\\ Using Social Media to Detect \\Urban Emergencies}
\titlerunning{Civique}  
%
\author{Diptesh Kanojia$^*$ \inst{1,}\inst{2,}\inst{3} \and Vishwajeet Kumar\thanks{These two authors contributed equally} \inst{1,}\inst{2,}\inst{3} \and Krithi Ramamritham\inst{1}}

%
%
%
\authorrunning{Diptesh Kanojia et al.}   
%


\institute{IIT Bombay, India,\\
\and
Monash Univerity, Australia,\\
\and
IITB-Monash Research Academy, India\\
\email{{diptesh,vishwajeet,krithi}@cse.iitb.ac.in} \\
}

\maketitle              
 \index{Kanojia, Diptesh} 
 \index{Kumar, Vishwajeet}  
 \index{Ramamritham, Krithi}

\begin{abstract}        



We present the Civique system for emergency detection in urban areas by monitoring micro blogs like Tweets. The system detects emergency related events, and classifies them into appropriate categories like ``fire'', ``accident'', ``earthquake'', etc. We demonstrate our ideas by classifying Twitter posts in real time, visualizing the ongoing event on a map interface and alerting users  with options to contact relevant authorities, both online and offline. We evaluate our classifiers for both the steps, i.e., emergency detection and categorization, and obtain F-scores exceeding  $70\%$ and $90\%$, respectively. We demonstrate Civique using a web interface and on an Android application, in realtime, and show its use for both tweet detection and visualization.


\end{abstract}

\section{Introduction}
With the surge in the use of social media, micro-blogging sites like Twitter\footnote{http://www.twitter.com}, Facebook\footnote{http://www.facebook.com}, and Foursquare\footnote{http://foursquare.com} have become household words. Growing ubiquity of mobile phones in highly populated developing nations has spurred an exponential rise in social media usage. The heavy volume of social media posts tagged with users' location information on micro-blogging website Twitter presents a unique opportunity to scan these posts. These Short texts (e.g. "tweets") on social media contain information about various events happening around the globe, as people post about events and incidents alike. Conventional web outlets provide emergency phone numbers (i.e. $100$, $911$), etc., and are fast and accurate. Our system, on the other hand, connects  its users through a relatively newer platform i.e. social media, and provides an alternative to these conventional methods. In case of their failure or when such means are busy/occupied, an alternative could prove to be life saving.

These real life events are reported on Twitter with different perspectives, opinions, and sentiment. Every day, people discuss events thousands of times across social media sites. We would like to detect such events in case of an emergency. Some previous studies\cite{www10} investigate the use of features such as keywords in the tweet, number of words, and context to devise a classifier for event detection. \cite{survey} discusses various techniques researchers have used previously to detect events from Twitter. \cite{www11} describe a system to automatically detect events about known entities from Twitter. This work is highly specific to detection of events only related to known entities. \cite{edbtmanoj16} discuss a system that returns a ranked list of relevant events given a user query. 

Several research efforts have focused on identifying events in real time(\cite{mdmkdd10}\cite{cucs} \cite{vldb13}\cite{www10}). These include  systems to detect emergent topics from Twitter in real time (\cite{mdmkdd10} \cite{alvanaki2011enblogue}), an online clustering technique for identifying tweets in real time \cite{cucs}, a system to detect localized events and also track evolution of such events over a period of time \cite{vldb13}. Our focus is on detecting urban emergencies as events from  Twitter messages. We classify events ranging from natural disasters to fire break outs, and accidents. Our system detects whether a tweet, which contains a keyword from a pre-decided list, is related to an actual emergency or not. It also classifies the event into its appropriate category, and visualizes the possible location of the emergency event on the map. We also support notifications to our users, containing the contacts of specifically concerned authorities, as per the category of their tweet. 

The rest of the paper is as follows: Section \ref{sec:motivation} provides the motivation for our work, and the challenges in building such a system. Section \ref{sec:approach} describes the step by step details of our work, and its results. We evaluate our system and present the results in Section \ref{sec:evaluation}. Section \ref{sec:description} showcases our demonstrations in detail, and Section \ref{sec:conclusion} concludes the paper by briefly describing the overall contribution, implementation and demonstration.

\section{Motivation and Challenges}
\label{sec:motivation}
In $2015$, $53\%$ of all unnatural deaths in India were caused by accidents, and $6\%$ by accidental fires\footnote{http://www.indiaspend.com/cover-story/fire-accidents-kill-54-every-day-yet-deaths-have-declined-81400}. Moreover, the Indian subcontinent suffered seven earthquakes in 2015, with the recent Nepal earthquake alone killing more than 9000 people and injuring $23,000$\footnote{https://en.wikipedia.org/wiki/April\_2015\_Nepal\_earthquake}. 
We believe  we can harness the current social media activity on the web to minimize  losses by quickly connecting affected people and the concerned authorities.  Our work is motivated by the following factors,
(a) Social media is very accessible in the current scenario. (The ``Digital India'' initiative by the Government of India promotes internet activity, and thus a pro-active social media.)
(b) As per the Internet trends  reported in $2014$\footnote{``Internet trends $2014$ report" by Mary Meeker, Kleiner Perkins Caufield \& Byers (KPCB)}, about 117 million Indians are connected to the Internet through mobile devices.
(c) A system such as ours can point out or visualize the affected areas precisely and help inform the authorities in a timely fashion.
(d) Such a system can be used on a global scale to reduce the effect of natural calamities and prevent loss of life.

There are several challenges in building such an application:
(a) Such a system expects a tweet to be location tagged. Otherwise, event detection techniques to extract the spatio-temporal data from the tweet can be vague, and lead to false alarms.
(b) Such a system should also be able to verify the user's credibility as pranksters may raise false alarms.
(c) Tweets are usually written in a very informal language, which requires a sophisticated language processing component to sanitize the tweet input before event detection.
(d) A channel with the concerned authorities should be established for them to take serious action, on alarms raised by such a system.
(e) An urban emergency such as a natural disaster could affect communications severely, in case of an earthquake or a cyclone, communications channels like Internet connectivity may get disrupted easily. In such cases, our system may not be of help, as it requires the user to be connected to the internet.
We address the above challenges and present our approach in the next section.

\section{Our Approach}
\label{sec:approach}
\begin{figure}[t]
\begin{center}
\includegraphics[scale=0.39]{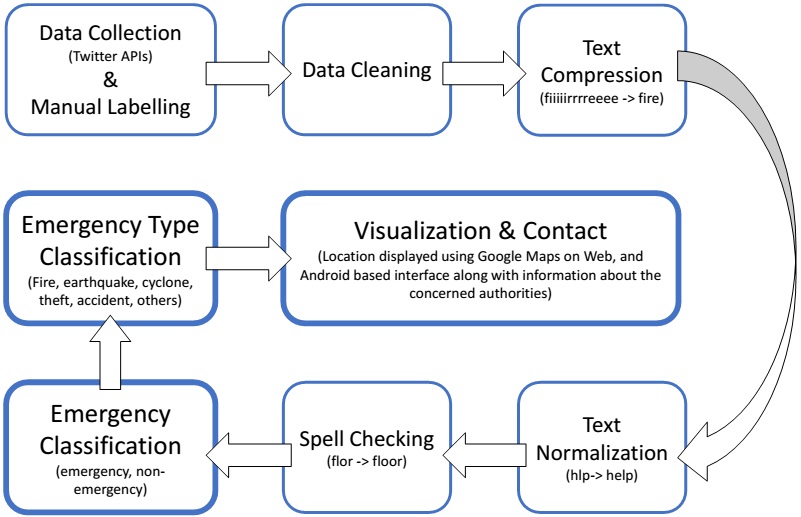}
\vspace{-0.50em}
\caption{System Architecture}
\vspace{-0.60em}
\label{fig:Arch1}
\end{center}
\end{figure}

\begin{table}[t]
\centering
\begin{tabular}{|l|l|}
\hline
\textbf{Input}  & @user heellllllppp!!! Firrreee at powai, lake lucene bldng 4th flor, hlp! \#fire \\ \hline
\textbf{Output} & heellllllppp Firrreee at powai , lake lucene bldng 4th flor , hlp \\ \hline
\end{tabular}
\caption{Cleaning Script Output}
\label{tab:sampleCleaning}
\end{table}
We propose a software architecture for Emergency detection and visualization as shown in figure \ref{fig:Arch1}. We collect data using Twitter API\footnote{https://dev.twitter.com/overview/documentation}, and perform language pre-processing before applying a classification model.  Tweets are labelled manually with \textless emergency\textgreater and \textless non-emergency\textgreater  labels, and later classified manually to provide labels according to the type of emergency they indicate. We use the manually labeled data for training our classifiers. 

We use traditional classification techniques such as Support Vector Machines(SVM), and Naive Bayes(NB) for training, and perform $10$-fold cross validation to obtain f-scores. Later,  in real time, our system uses the Twitter streaming APIs to get data, pre-processes it using the same modules, and detects emergencies using the classifiers built above. The tweets related to emergencies are displayed on the web interface along with the location and information for the concerned authorities. The pre-processing of Twitter data obtained is needed as it usually contains ad-hoc abbreviations, phonetic substitutions, URLs, hashtags, and a lot of misspelled words. We use the following language processing modules for such corrections.

\subsection{Pre-Processing Modules}
\label{sec:preprocessing}
We implement a cleaning module to automate the cleaning of tweets obtained from the Twitter API. We remove URLs, special symbols like @ along with the user mentions, Hashtags and any associated text. We also replace special symbols by blank spaces, and inculcate the module as shown in figure \ref{fig:Arch1}.

An example of such a sample tweet cleaning is shown in table \ref{tab:sampleCleaning}.

While tweeting, users often express their emotions by stressing over a few characters in the word.  For example, usage of words like \emph{hellpppp, fiiiiiireeee, ruuuuunnnnn, druuuuuunnnkkk, soooooooo} actually corresponds to \emph{help, fire, run, drunk, so} etc. We use the compression module implemented by \cite{kanojia2015transchat} for converting terms like ``pleeeeeeeaaaaaassseeee'' to ``please''.

It is unlikely for an English word to contain the same character consecutively for three or more times. We, hence, compress all the repeated windows of character length greater than two, to two characters. For example ``pleeeeeaaaassee'' is converted to ``pleeaassee''. Each window now contains two characters of the same alphabet in cases of repetition. Let \emph{n} be the number of windows, obtained from the previous step. We, then, apply brute force search over $2^{n}$ possibilities to select a valid dictionary word.

Table \ref{tab:sampleCompression} contains sanitized sample output from our compression module for further processing.

\begin{table}[ht]
\footnotesize
\centering
\begin{tabular}{|c|c|}
\hline
                   \textbf{Sample Input}        & \textbf{Sample Output} \\ \hline
fiiiirreeeeee & fire           \\ \hline
helllllp m sttuuuuuccckk          & help m stuck           \\ \hline
\end{tabular}
\caption{Sample output of Compression module}
\label{tab:sampleCompression}
\vspace{-0.90em}
\end{table}

Text Normalization is the process of translating ad-hoc abbreviations, typographical errors, phonetic substitution and ungrammatical structures used in text messaging (Tweets and SMS) to plain English. Use of such language (often referred as Chatting Language) induces noise which poses additional processing challenges.

We use the normalization module implemented by \cite{kanojia2015transchat} for text normalization. Training process requires a Language Model of the target language and a parallel corpora containing aligned un-normalized and normalized word pairs. Our language model consists of $15000$ English words taken from various sources on the web.

Parallel corpora was collected from the following sources:
\begin{enumerate}
\item Stanford Normalization Corpora which consists of $9122$ pairs of un-normalized and normalized words / phrases.

\item The above corpora, however, lacked acronyms and short hand texts like \textit{2mrw, l8r, b4, hlp, flor} which are frequently used in chatting. We collected $215$ pairs un-normalized to normalized word/phrase mappings via crowd-sourcing.

\end{enumerate}

Table \ref{tab:sampleNormalization} contains input and normalized output from our module.
\begin{table}[t]
\footnotesize
\centering
\begin{tabular}{|c|c|}
\hline
                   \textbf{Input Sentence}        & \textbf{Output Sentence} \\ \hline
bldng $4th$ flor, hlp & building $4th$ flor, help          \\ \hline
\end{tabular}
\caption{Sample output of Normalization module}
\label{tab:sampleNormalization}
\vspace{-0.60em}
\end{table}

Users often make spelling mistakes while tweeting. A spell checker makes sure that a valid English word is sent to the classification system. We take this problem into account by introducing a spell checker as a \emph{pre-processing} module by using the JAVA API of Jazzy spell checker\footnote{http://sourceforge.net/projects/jazzy/} for handling spelling mistakes.
\\
\\An example of correction provided by the Spell Checker module is given below:-
\\
\\\textit{\textbf{Input:} building $4th$ \textbf{flor}, help}
\\\textit{\textbf{Output:} building $4th$ \textbf{floor}, help}
\\

Please note that, our current system performs compression, normalization and spell-checking if the language used is English. The classifier training and detection process are described below.

\subsection{Emergency Classification}
\label{sec:emergencyClassification}
The first classifier model acts as a filter for the second stage of classification. We use both SVM and NB to compare the results and choose SVM later for stage one classification model, owing to a better F-score. The training is performed on tweets labeled with classes <emergency>, and <non-emergency> based on \textit{unigrams} as features. We create word vectors of strings in the tweet using a filter available in the WEKA API\cite{hall2009weka}, and perform cross validation using standard classification techniques.

\subsection{Type Classification}
\label{sec:typeClassification}
We employ a multi-class Naive Bayes classifier as the second stage classification mechanism,  for categorizing tweets appropriately, depending on the type of emergencies they indicate. This multi-class classifier is trained on data manually labeled with classes. We tokenize the training data using ``NgramTokenizer'' and then, apply a filter to create word vectors of strings before training. We use ``trigrams'' as features to build a model which, later, classifies tweets into appropriate categories, in real time. We then perform cross validation  using standard techniques to calculate the results, which are shown under the label ``Stage 2'', in table \ref{tab:classResults}.

\begin{table}[t]
\footnotesize
\centering
\begin{tabular}{|c|c|c|}
\hline \textbf{Classifier} & \textbf{Stage 1}        & \textbf{Stage 2} \\ \hline
\textbf{SVM} & \textbf{72.5\%} & $90.5\%$ \\ \hline
\textbf{NB} & $67.9\%$  & \textbf{92\%} \\ \hline
\end{tabular}
\caption{Classification Results}
\label{tab:classResults}
\vspace{-0.30em}
\end{table}
\subsection{Location Visualizer}
\label{sec:visualizer}
We use Google Maps Geocoding API\footnote{https://developers.google.com/maps/documentation/geocoding} to display the possible location of the tweet origin based on longitude and latitude. Our visualizer presents the user with a map and pinpoints the location with custom icons for earthquake, cyclone, fire accident etc. Since we currently collect tweets with a location filter for the city of "Mumbai", we display its map location on the interface. The possible occurrences of such incidents are displayed on the map as soon as our system is able to detect it.

We also display the same on an Android device using the WebView functionality available to developers, thus solving the issue of portability. Our system displays visualization of the various emergencies detected on both web browsers and mobile devices.

\section{Evaluation}
\label{sec:evaluation}
We evaluate our system using automated, and manual evaluation techniques. We perform 10-fold cross validation to obtain the F-scores for our classification systems. We use the following technique for dataset creation. We test the system in realtime environments, and tweet about fires at random locations in our city, using test accounts. Our system was able to detect such tweets and detect them with locations shown on the map.

\subsection{Dataset Creation}
\label{sec:dataCreation}
We collect data by using the Twitter API for saved data, available for public use. For our experiments we collect 3200 tweets filtered by keywords like ``fire'', ``earthquake'', ``theft'', ``robbery'', ``drunk driving'', ``drunk driving accident'' etc. Later, we manually label tweets with \textless emergency\textgreater  and \textless non-emergency\textgreater labels for classification as stage one. Our dataset contains $1313$ tweet with \textit{positive} label \textless emergency\textgreater   and $1887$ tweets with a \textit{negative} label \textless non-emergency\textgreater. We create another dataset with the positively labeled tweets and provide them with category labels like ``fire'', ``accident'', ``earthquake'' etc.

\begin{figure}[h]
\begin{center}
\includegraphics[scale=0.30]{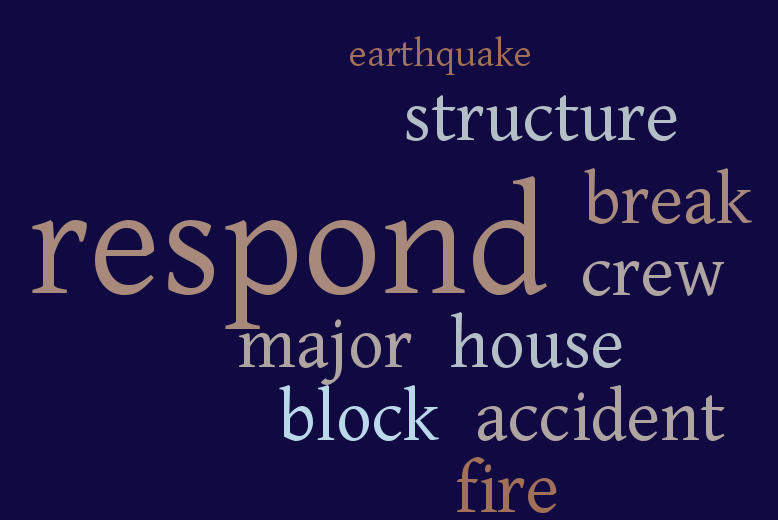}
\caption{Word Cloud of top attributes}
\vspace{-0.60em}
\label{fig:wordCloud}
\end{center}
\end{figure}
\vspace{-1.80em}

\subsection{Classifier Evaluation}
\label{sec:classifyEval}
The results of 10-fold cross-validation performed for stage one are shown in table \ref{tab:classResults}, under the label ``Stage 1''. In table \ref{tab:classResults}, For ``Stage 1'' of classification, F-score obtained using SVM classifier is $72.5\%$ as shown in row 2, column 2. We also provide the system with sample tweets in real time and assess its ability to detect the emergency, and classify it accordingly. The classification training for Stage 1 was performed using two traditional classification techniques SVM and NB. SVM outperformed NB by around $4\%$ and became the choice of classification technique for stage one.

Some false positives obtained during manual evaluation are, ``I am sooooo so drunk right nowwwwwwww'' and ``fire in my office , the boss is angry''. These occurrences show the need of more labeled gold data for our classifiers, and some other features, like Part-of-Speech tags, Named Entity recognition, Bigrams, Trigrams etc. to perform better.

The results of 10-fold cross-validation performed for stage two classfication model are also shown in table \ref{tab:classResults}, under the label ``Stage 2''. The training for stage two was also performed using both SVM and NB, but NB outperformed SVM by around $1\%$ to become a choice for stage two classification model.

We also perform attribute evaluation for the classification model, and create a word cloud based on the output values, shown in figure \ref{fig:wordCloud}. It shows that our classifier model is trained on appropriate words, which are very close to the emergency situations viz. ``fire'', ``earthquake'', ``accident'', ``break'' (Unigram representation here, but possibly occurs in a bigram phrase with ``fire'') etc. In figure \ref{fig:wordCloud}, the word cloud represents the word ``respond'' as the most frequently occurring word as people need urgent help, and quick response from the assistance teams.

\section{Demostration Description}
\label{sec:description}

\begin{figure}[t]
\begin{center}
\includegraphics[scale=0.35]{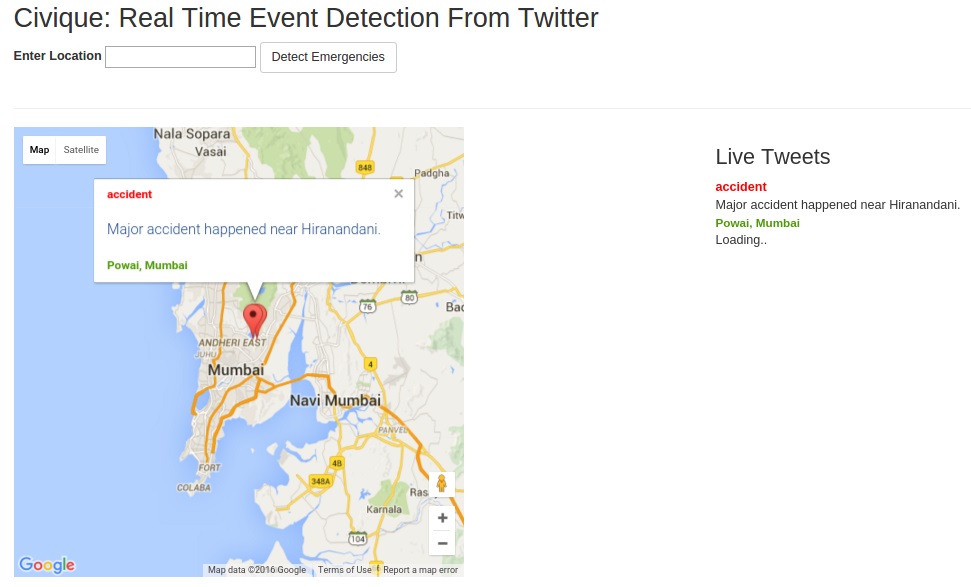}
\vspace{-0.50em}
\caption{Screenshot: Web Interface}
\vspace{-0.60em}
\label{fig:web}
\end{center}
\end{figure}

\begin{figure}[t]
	\begin{floatrow}
    \ffigbox{\includegraphics[scale = 0.10]{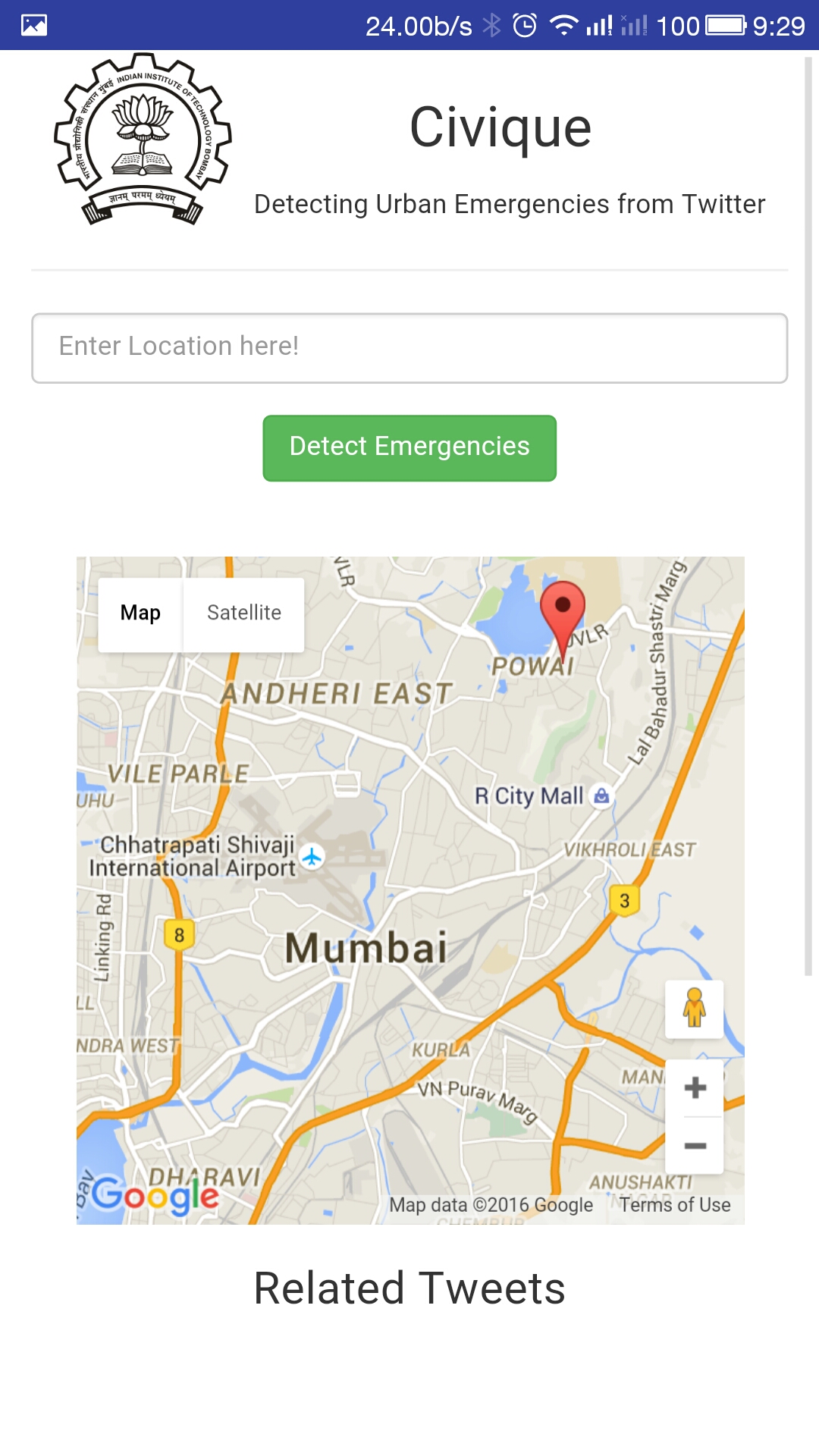}}
    {
    \caption{Screenshot: Mobile Interface}
    \label{fig:and1}
    }
    \ffigbox{\includegraphics[scale = 0.10]{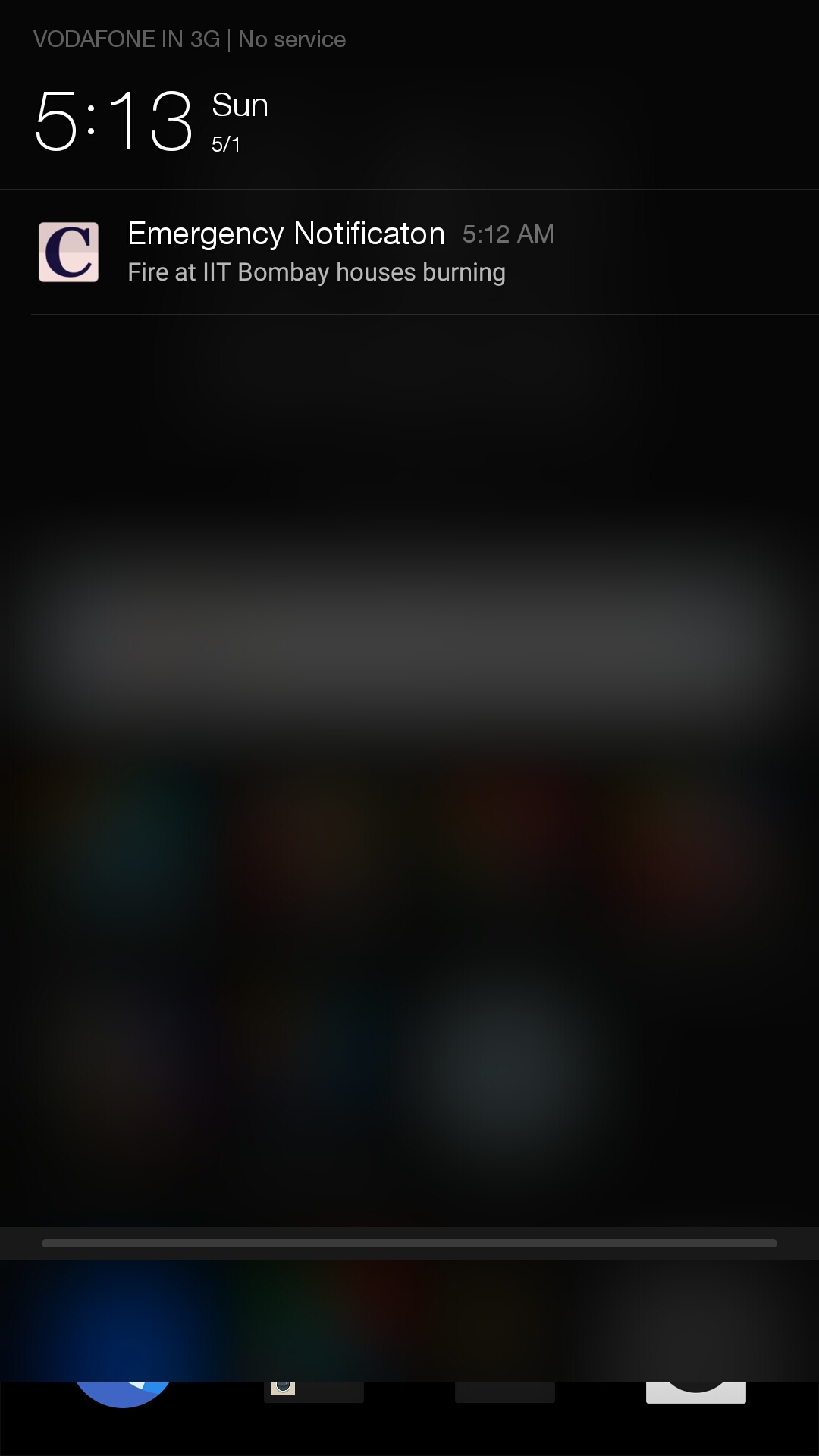}}		
    {
    \caption{Screenshot: Generated Notification}      		
    \label{fig:and2}
	}
    \end{floatrow}
\end{figure}
        
Users interact with Civique through its Web-based user interface and Android based application interface. The features underlying Civique are demonstrated through the following two show cases:\\
\\
\textbf{Show case 1: Tweet Detection and Classification}

This showcase aims at detecting related tweets, and classifying them into appropriate categories. For this, we have created a list of filter words, which are used to filter tweets from the Twitter streaming API. These set of words help us  filter the tweets related to any  incident. We will tweet, and users are able to see how our system captures such tweets and classifies them. Users should be able to see the tweet emerge as an incident on the web-interface, as shown in figure \ref{fig:web} and the on the android application, as shown in figure \ref{fig:and1}. Figure \ref{fig:and2} demonstrates how a notification is generated when our system detects an emergency tweet. When a user clicks the emerged spot, the system should be able to display  the sanitized version / extracted spatio-temporal data from the tweet. We test the system in a realtime environment, and validate our experiments. We also report the false positives generated during the process in section \ref{sec:classifyEval} above.\\
\\
\textbf{Show case 2: User Notification and Contact Info.}

Civique includes a set of local contacts for civic authorities who are to be / who can be contacted in case of various emergencies. Users can see how Civique detects an emergency and classifies it. They can also watch how the system generates a notification on the web interface and the Android interface, requesting them to contact the authorities for emergencies. Users can change their preferences on the mobile device anytime and can also opt not to receive  notifications. Users should be able to contact the authorities online using the application, but in case the online contact is not responsive, or in case of a sudden loss of connectivity, we provide the user with the offline contact information of the concerned civic authorities along with the notifications.

\section{Conclusions}
\label{sec:conclusion}

Civique is a system which detects urban emergencies like earthquakes, cyclones, fire break out, accidents etc. and visualizes them on both on a browsable web interface and an Android application. We collect data from the popular micro-blogging site Twitter and use language processing modules to sanitize the input. We use this data as input to train a two step classification system, which indicates whether a tweet is related to an emergency or not, and if it is, then what category of emergency  it belongs to. We display such positively classified tweets along with their type and location on a Google map, and notify our users to inform the concerned authorities, and possibly evacuate the area, if his location matches the affected area. We believe such a system can help the disaster management machinery, and government bodies like Fire department, Police department, etc., to act swiftly, thus minimizing the loss of life.

Twitter users use slang, profanity, misspellings and neologisms. We, use standard cleaning methods, and combine NLP with Machine Learning (ML) to further our cause of tweet classification. At the current stage, we also have an Android application ready for our system, which shows the improvised, mobile-viewable web interface.

In the future, we aim to develop detection of emergency categories on the fly, obscure emergencies like ``airplane hijacking''  should also be detected by our system. We plan to analyze the temporal sequence of the tweet set from a single location to determine whether multiple problems on the same location are the result of a single event, or relate to multiple events.

\bibliographystyle{splncs03}
\bibliography{vldb_sample}

\begin{thebibliography}{10}
\providecommand{\url}[1]{\texttt{#1}}
\providecommand{\urlprefix}{URL }

\bibitem{vldb13}
Abdelhaq, H., Sengstock, C., Gertz, M.: In: EvenTweet: online localized event
  detection from twitter (2013)

\bibitem{edbtmanoj16}
Agarwal, M.K., Bansal, D., Garg, M., Ramamritham1, K.: Keyword search on
  microblog data streams:finding contextual messages in real time. In:
  Proceedings of 19th International Conference on Extending Database Technology
  (EDBT). pp. 15--18 (2016)

\bibitem{alvanaki2011enblogue}
Alvanaki, F., Sebastian, M., Ramamritham, K., Weikum, G.: Enblogue: emergent
  topic detection in web 2.0 streams. In: Proceedings of the 2011 ACM SIGMOD
  International Conference on Management of data. pp. 1271--1274. ACM (2011)

\bibitem{survey}
atefeh, F.: A survey of techniques for event detection in twitter. In:
  computational intelligence

\bibitem{mdmkdd10}
Cataldi, M., caro, L.D., schifanella, C.: Emerging topic detection on twitter
  based on temporal and social terms evaluation. In: Proceedings of the Tenth
  International Workshop on Multimedia Data Mining (2010)

\bibitem{hall2009weka}
Hall, M., Frank, E., Holmes, G., Pfahringer, B., Reutemann, P., Witten, I.H.:
  The weka data mining software: an update. ACM SIGKDD explorations newsletter
  11(1),  10--18 (2009)

\bibitem{cucs}
Hila, B., Mor, N., Luis, G.: In: Beyond Trending Topics: Real-World Event
  Identification on Twitter (2011)

\bibitem{kanojia2015transchat}
Kanojia, D., Dhuliawala, S., Gupta, N., Mishra, A., Bhattarcharyya, P.:
  Transchat: Cross-lingual instant messaging for indian languages. In: Twelth
  International Conference on Natural Language Processing. ICON (2015)

\bibitem{www11}
Popescu, A.M., Pennacchiotti, M., Paranjpe, D.: Extracting events and event
  descriptions from twitter. In: Proceedings of the 20th international
  conference companion on World wide web (2011)

\bibitem{www10}
Sakaki, T., Okazaki, M., Matsuo, Y.: In: Earthquake shakes Twitter users:
  real-time event detection by social sensors (2010)

\end{thebibliography}

\end{document}